\documentclass[runningheads]{llncs}

\usepackage[T1]{fontenc}    
\usepackage{hyperref}       
\usepackage{url}            
\usepackage{booktabs}       
\usepackage{amsmath}
\usepackage{amsfonts}       
\usepackage{easyeqn}
\usepackage{xcolor}
\usepackage{graphicx}

\usepackage[linesnumbered,ruled]{algorithm2e}
\SetKwComment{Comment}{\# }{}

\newcommand{\reals}{\mathbb{R}}

\def\xm{\mathbf{X}} 
\def\xv{\mathbf{x}} 

\def\wp{\mathbf{w}}
\def\wpe{\widehat{\wp}}

\def\veps{\boldsymbol{\varepsilon}} 

\newcommand\CM{\boldsymbol{\Sigma}} 

\def\sighat{\widehat{\sigma}} 

\newcommand\what{\widehat{w}} 

\newcommand\DS{\mathcal{D}} 

\begin{document}

\title{Scalable computation of prediction intervals for neural networks via matrix sketching}

\author{Alexander Fishkov \and Maxim Panov}

\institute{
Skolkovo Institute of Science and Technology\\
Bolshoy Boulevard 30, bld. 1\\
Moscow, Russia 121205\\
\email{\{alexander.fishkov, m.panov\}@skoltech.ru}, 
}

\maketitle

\begin{abstract}
    Accounting for the uncertainty in the predictions of modern neural networks is a challenging and important task in many domains. Existing algorithms for uncertainty estimation require modifying the model architecture and training procedure (e.g., Bayesian neural networks) or dramatically increase the computational cost of predictions such as approaches based on ensembling. This work proposes a new algorithm that can be applied to a given trained neural network and produces approximate prediction intervals. The method is based on the classical delta method in statistics but achieves computational efficiency by using matrix sketching to approximate the Jacobian matrix. The resulting algorithm is competitive with state-of-the-art approaches for constructing predictive intervals on various regression datasets from the UCI repository.

\end{abstract}

\section{Introduction and related work}
\label{sec:intro}
  Modern neural networks achieve great results on various predictive modelling tasks. In the fields like computer vision and natural language processing neural-network based approaches are ubiquitous. When the task at hand has high cost of wrong decisions one has to assess the reliability of the model's predictions. There exist different ways to proceed with this task for neural networks. 

  The standard approach for uncertainty estimation is ensembling which proposes to train several neural networks from different starting parameter values. Next, one can compute uncertainty estimates based on the discrepancies between the predictions of ensemble members~\cite{heskes1997practical,deepens2016}. Ensembling usually gives high quality uncertainty estimates but for the price of large computational overhead both on training and inference stages. The different direction is to make weights of the network stochastic via Bayesian approach which results in the concept of Bayesian neural networks~\cite{jospin2020hands}. One can then quantify predictive uncertainty using the posterior distribution of the weights conditioned on the training dataset. Bayesian inference is intractable for practical neural networks and thus approximate methods like variational inference~\cite{graves2011practical,blundell2015weight,rezende2015variational} and stochastic gradient MCMC~\cite{chen2014sghmc} are in use. These methods have a higher computational cost compared to training regular neural networks which limits their practical applicability. The other popular alternative is dropout which adds stochasticity to a standard neural network via randomly setting some of the weights to zero. This technique leads to the regularization of training~\cite{srivastava2014dropout} and can provide uncertainty estimates if applied at prediction time~\cite{gal2016dropout,Tsymbalov2018,tsymbalov2020dropout,shelmanov-etal-2021-certain}.

  All the mentioned methods require either a modification of network architecture or training multiple models. As there are many existing pre-trained models for various tasks, it is beneficial to have a method that can be applied to a given neural network in order to obtain high-quality uncertainty estimates.
  
  In this work we will focus on the classical form of uncertainty estimates: prediction intervals~\cite{casella2002statistical}. We propose a general method to construct approximate prediction intervals for a trained neural network model in the non-linear regression problem. Our method is based on the classical ideas of linearization (or first-order Taylor expansion) and so-called \textit{delta method} (see e.g.~\cite{doob1935}). While the original method requires inverting the estimated parameter covariance matrix of the model, we propose an efficient computational scheme based on SVD and matrix sketching techniques. These techniques make the proposed \textit{Important Directions} method viable for modern deep neural networks. 

  The rest of the work is organized as follows. Section~\ref{sec:algo} is devoted to the formal problem statement and introduces the proposed method.  Section~\ref{sec:experiments} describes the implementation and experiments. After that Section~\ref{sec:discussion} points to possible extensions of the considered method and Section~\ref{sec:conclusions} concludes the work.

\section{Approximate prediction intervals for neural networks}
\label{sec:algo}
\subsection{Prediction intervals for neural networks}

  We will be working in a general regression setting, where we are given a training dataset $\DS=\{\xv_i, y_i\}_{i=1}^{n}, \: \xv_i \in \reals^m, \: y_i \in \reals$. The task is to estimate the hypothesised relationship $y_i \approx f(\xv_i, \wp)$ by providing an estimate $\widehat{\wp}$ of the parameter vector $\wp \in \reals^p$. To quantify the approximate nature of the statement, an unobserved noise random variable is introduced:
  \begin{EQA}[c]
  \label{eqn:nonlin_reg}
    y_i = f(\xv_i, \wp) + \veps_i, \quad \veps_i \sim \mathcal{N}(0, \sigma^2), \quad i = 1, \dots, n. 
  \end{EQA}
  A level $(1 - \alpha)$ prediction interval for the function value $y_0 = f(\xv_0, \wp)$ is a random interval $\left[L_\alpha(\xv_0, \DS), U_\alpha(\xv_0, \DS)\right]$ with the following property:
  \begin{EQA}[c]
    \mathbb{P}\Bigl( y_0 \in \bigl[L_\alpha(\xv_0, \DS), U_\alpha(\xv_0, \DS)\bigr] \Bigr) \geq 1 - \alpha.
  \end{EQA}

\noindent \textbf{Linear regression.}
  In the simplest case of a linear relationship $f$ there is a closed-form expression for the prediction interval. For ease of exposition need to slightly modify the training set here $\xv_i \in \reals^{m+1}$ with a dummy first dimension equal to 1 and $\xm$ is a matrix with the vectors $\xv_i$ as rows. Using the fact that the noise is normally distributed, the following formula for interval endpoints can be derived~\cite{seber_lee2003linear}:
  \begin{EQA}[c]
    \widehat{y}_0 \: \pm \: t_{n-m-1}^{(\alpha/2)} \sighat \sqrt{1 + \xv_0^T \left( \xm^T \xm \right)^{-1}  \xv_0},
  \end{EQA}
  where $\sighat$ is an estimate of the noise standard deviation, $\widehat{y}_0$ is a linear regression prediction at point $\xv_0$ and $t_{n-m-1}^{(\alpha/2)}$ is a level $\alpha/2$ quantile of the Student's $t$-distribution with $n-m-1$ degrees of freedom.

  From the expression under the square root we see that even in this simple case the required computation scales quadratically with the size of the whole dataset and cubically with the feature dimension. Overall, computational complexity is $O(n^2m+m^3)$.

\subsubsection{Nonlinear regression.}
  Now we move to the general setting of~\eqref{eqn:nonlin_reg} with $f$ being a neural network with parameters $\wp$. In~\cite{Veaux1998PI} authors provide a method of constructing a prediction interval for a trained neural network based on delta-method. We briefly state main results here while the derivation and discussion can be found in the original paper.

  Let us assume that parameter estimates $\wpe$ for the model are obtained by minimizing sum-of-squares loss with $L_2$ regularization with parameter $\lambda$ based on the training dataset $\DS=\{\xv_i, y_i\}_{i=1}^{n}$. We introduce some additional notations. Matrix $J$ consists of the gradients of the output of the neural network, computed at the final parameter estimates $\wpe$:
  \begin{EQA}[c]
    J = \{J_{ij}\}_{i, j= 1}^{n, m}, \; \; J_{i,j} = \left.\frac{\partial f}{\partial w_j} \right\vert_{\xv=\xv_i, \wp=\wpe},
  \end{EQA}
  and matrix $\CM$ is the approximate parameter covariance matrix:
  \begin{EQA}[c]
  \label{eq:1}
    \CM^{-1} = (J^T J + \lambda I)^{-1} (J^T J) (J^T J + \lambda I)^{-1}.
  \end{EQA}

  Let $\xv_0$ be a new unseen point where we aim to predict the response value $y_0$. Additionally, let $g_0 = \left.\frac{\partial f}{\partial w_j} \right\vert_{\xv=\xv_0, \wp=\wpe}$ be the gradient of the output of the neural network at the new point $\xv_0$.
  The approximate level $(1-\alpha)$ prediction interval at a new point $\xv_0$ is given by~\cite{Veaux1998PI}:
  \begin{EQA}[c]
    \widehat{y}_0 \pm t^{\alpha/2}_{n-p^*} \widehat{s} \sqrt{1 + g_0^T \CM^{-1} g_0}. \label{eq2}
  \end{EQA}
  Here, $t^{\alpha/2}_{n-p^*}$ is the $\alpha/2$-level quantile of Student's \textit{t}-distribution with $n-p^*$ degrees of freedom. Quantity $p^*$ is the so-called \textit{effective number of parameters}. Authors of~\cite{Veaux1998PI} propose the following approximation for it:
  \begin{EQA}[c]
    p^* = \mathrm{Tr}(2 H - H^2) ~~ \text{with} ~~ H = J(J^T J + \lambda I)^{-1} J^T.
  \end{EQA}
  The last component we have to cover is $\widehat{s}$, which is an estimate of the noise variance $\sigma$. This quantity can be found from the sum of squared residuals of the model on the training data:
  \begin{EQA}[c]
  \label{eqn:s_hat}
    \widehat{s}^2 = \frac{\sum_{i=1}^{n} (y_i - \widehat{y}_i)^2}{n - p^*} = \frac{\sum_{i=1}^{n} \bigl(y_i - f(\xv_i, \wpe)\bigr)^2}{n - p^*}.
  \end{EQA}

  From the applied point of view, we can divide this method into two parts: calculating $\CM$ and calculating $p^*$. Computational complexity of both parts is $O(n m^2 + m^3)$, not including the cost of computing gradients. Space complexity is at least $O(nm+m^2)$ since we need to store the full covariance matrix and the matrix of gradients.

  Unfortunately, for most modern deep neural networks this approach can not be applied as is. For example, in contemporary computer vision, a neural network can have tens of millions of parameters ($m$) and millions of examples in the training set ($n$). This will make the matrices involved in the formulas above prohibitively large to carry out computations.

  In this work we propose a set of approximation techniques to reduce the computational burden of the formulas above and obtain a practical algorithm to construct approximate prediction intervals for deep neural networks.

\subsection{Simplification using SVD}
  If we assume that matrix $J$ admits  the following singular value decomposition:
  \begin{EQA}[c]
    J = UDV^T, ~~ D = \mathrm{diag} (d_1, d_2, \dots, d_m),
  \label{eq_svd}
  \end{EQA}
  we can then simplify formula~\eqref{eq:1} using the fact that $U$ and $V$ are unitary matrices:
  \begin{EQA}
    && J^TJ = VDU^T UDV^T = VD^2V^T, \\
    && (J^TJ + \lambda I)^{-1} = (VD^2V^T + \lambda I)^{-1} = (V(D^2 + \lambda I)V^T)^{-1} = V(D^2 + \lambda I)^{-1}V^T
  \end{EQA}
  and obtain
  \begin{EQA}[c]
    \CM^{-1} = V(D^2 + \lambda I)^{-1}V^T VD^2V^T V(D^2 + \lambda I)^{-1}V^T = VD^2(D^2 + \lambda I)^{-2}V^T.
  \end{EQA}
  We can also observe that
  \begin{EQA}
  \label{eqn:d_sigma}
    D^2(D^2 + \lambda I)^{-2} & = & \mathrm{diag}\left( \left\{\frac{d_j^2}{(d_j^2 + \lambda)^2} \right\}_{j=1}^{m}\right).
  \end{EQA}
  Using this form we can greatly simplify the computation of the formula~\eqref{eq:1} if the SVD decomposition is precomputed.

\subsection{Proposed approximation}
  While obtaining the full decomposition~\eqref{eq_svd} may still be intractable, we can use a truncated rank-$k$ SVD. However, most existing implementations assume that the target matrix is stored and accessed in a sparse format.

  Since $J$ is a matrix of gradients of a neural network, it is dense, and, thus, we have a slightly different setup. First, the target matrix is accessible row-by-row, because modern automatic differentiation packages allow computation of the gradients of a function that outputs a single scalar. Second, we can only store a limited number of rows because of the large number of trainable parameters in the neural network. 
  
  Based on these two requirements, we need an online algorithm that computes a low rank approximation of a matrix by reading through its rows one at a time. Low rank matrix approximation in the presence of constraints on access to the matrix elements is sometimes called matrix sketching. We refer to~\cite{Sarkar2020SpectralAF} and~\cite{Liberty2013SimpleAD} for further reading. 

  In this work, we consider an online low rank matrix approximation algorithm Robust Frequent Directions (RFD)~\cite{robust_dirs}. RFD is a modification of Frequent Directions (FD) algorithm~\cite{freq_dirs,Liberty2013SimpleAD}. Given a matrix $A$, the algorithm approximates $A^TA$ by $B^TB + \lambda' I$, where matrix $B$ has only $k<<n$ rows. Reading matrix $A$ row-by-row, RFD updates the approximation $B$ and the regularization parameter $\lambda'$ simultaneously\footnote{When $A^TA + \lambda I$ is the target matrix the final approximation becomes $B^TB + (\lambda + \lambda') I$}.

  \begin{algorithm}[t]
    \label{alg:id_inner}
    \DontPrintSemicolon
    \SetKwInOut{Input}{Input}
    \SetKwInOut{Output}{Output}
    \SetKwFunction{SVD}{SVD}
    \caption{Update Approximation}
    \SetAlgoLined

    \Input{$B \in \reals^{2k \times m}$ -- current approximation to the target matrix, $\lambda$ -- regularization parameter used for network training, $\lambda'$ -- current additional regularization parameter, $R$ -- next row of the target matrix }

    \Output{$B'$ -- updated low-rank approximation, $\lambda''$ -- updated additional regularization coefficient}

    $r \leftarrow \textrm{index of the first zero row of } B$ \;
    $B_r \leftarrow R$ \;
    \uIf{$r = 2k$}{
        $U, D, V$ $\leftarrow$ \SVD{$B$} \;
        $S, \textrm{idx} \leftarrow$ top $k$ entries of $D$ based on $score(D, \lambda)$ together with indices\;
        $\delta \leftarrow \min S$ \;
        $B' \leftarrow 0^{2k \times m}$ \;
        $B'_{1:k} \leftarrow \sqrt{\max (S^2-\delta^2 I, 0)} V_{\textrm{idx}}^T$  \Comment{use singular vectors indexed with \textrm{idx} and scale corresponding singular values}
        $\lambda'' \leftarrow \lambda' + \delta^2 / 2$ \;
    }
    \uElse{
        $B' \leftarrow B$ \;
        $\lambda'' \leftarrow \lambda'$ \;
    }
    \Return {$B', \; \lambda''$}
  \end{algorithm}

\subsubsection{Parameter covariance.}
  We propose a new algorithm to approximate formula~\eqref{eq:1} using a modification of the RFD that we call Robust Important Directions, presented in Algorithm 1. The key element of the algorithm is the use of a scoring function to select a pivotal singular value (line 6) instead of just using the minimal one. This scoring function is chosen based on a higher-level problem at hand. In our case the end goal is estimating the parameter covariance matrix and~\eqref{eqn:d_sigma} can be directly used here:
  \begin{EQA}[c]
  \label{eqn:score_func}
    score(D, \lambda) = \left[\frac{d_j^2}{(d_j^2 + \lambda)^2}\right]_{j=1}^{m}.
  \end{EQA}
  Since we want to preserve as much information about $\CM^{-1}$ as possible, this score function will help the algorithm retain the largest singular values.

  \begin{algorithm}[t]
    \label{alg:id_outer}
    \DontPrintSemicolon
    \SetKwInOut{Input}{Input}
    \SetKwInOut{Output}{Output}
    \SetKwFunction{UpdateApproximation}{UpdateApproximation}
    \caption{Robust Important Directions}
    \SetAlgoLined
    \Input{$k$ -- rank of the approximation, $A \in \reals^{n \times m}$ -- target matrix, $\lambda$ -- regularization parameter}
    \Output{$B$ -- low-rank approximation of $A$, $\lambda_n$ -- corrected regularization coefficient}
    $B \leftarrow 0^{2k \times m}$ \; 
    $\lambda_0 \leftarrow 0$ \;
    \ForEach {row $A_i \in A$}{
     $B, \lambda_i \leftarrow$ \UpdateApproximation{$B, \lambda, \lambda_{i-1}, A_i$} \; 
    }
    \Return {$B_{1:k}, \lambda_n$}
  \end{algorithm}

  After going through our training dataset one time we store the final SVD decomposition of the matrix $B$. This approximation of matrix $J$ is used later for computation of the intervals.

  Computational complexity of this procedure is the same as for Frequent Directions~\cite{freq_dirs}: $O(nmk)$, where $k$ is the rank of the approximation. We again ignore the cost of computing gradients since it is done only once and the resulting computational complexity will depend on the specific architecture. Space complexity is only $O(km)$ as we store $B \in \reals^{2k, m}$.

\subsubsection{Noise variance.}
  So far uncovered ingredient of the formula~\eqref{eq2} is the estimation of the noise variance $\widehat{s}^2$. Using our SVD decomposition of $J$ we can represent $H$ in the following form:
  \begin{EQA}[c]
    H = J(J^T J + \lambda I)^{-1} J = UDV^T V(D^2 + \lambda I)^{-1}V^T VDU^T = UD^2(D^2 + \lambda I)^{-1}U^T
  \end{EQA}
  and $D^2(D^2 + \lambda I)^{-1}$ can be computed by~\eqref{eqn:d_sigma}.

  To compute the unknown quantity $p^*$ we need to find the trace of the equation depending on $H$. Using the properties of SVD decomposition and trace we get:
  \begin{EQA}[c]
    p^* = \mathrm{Tr}(2 H - H^2) = 2\mathrm{Tr}(H) - \mathrm{Tr}(H^2)
    = \sum_{j=1}^m \frac{2d_j^2}{(d_j^2 + \lambda)} - \frac{d_j^4}{(d_j^2 + \lambda)^2}.
  \end{EQA}
  Having obtained this simple expression, we can evaluate it using our truncated SVD approximation.

\subsubsection{Algorithms.}  
  We collect the steps described above in the following subprograms. 
  
  Algorithm~\ref{alg:id_inner} computes an update to existing low rank approximation. It runs when a new row of the target matrix is available: if the buffer (matrix $B$) is full, we perform SVD and truncate our approximation. We use this later as a subroutine.
  
  Algorithm~\ref{alg:id_outer} is the procedure for computing low-rank approximation of a matrix for the case when a $score$ function is provided (implicitly used in Algorithm~\ref{alg:id_inner}). This algorithm can be used as a drop-in replacement for other low-rank approximation algorithms when more information about the problem is available. The end user will select the $score$ function based on the task at hand.
  
  Algorithm~\ref{alg:covar} summarizes all the steps that are required to compute prediction intervals of a trained neural network for the proposed \textit{Important Directions} approach. It precomputes a compact low rank approximation to the inverse covariance matrix of the network parameters and effective number of parameters. To construct the intervals for new data points the end user can employ equation~\eqref{eq2} together with these approximations.

  \begin{algorithm}[t]
    \label{alg:covar}
    \DontPrintSemicolon
    \SetKwInOut{Input}{Input}
    \SetKwInOut{Output}{Output}
    \SetKwFunction{UpdateApproximation}{UpdateApproximation}
    \SetKwFunction{SVD}{SVD}
    \SetAlgoLined
    \Input{$f(\xv, \wpe)$ - trained neural network, $\lambda$ - $L_2$-regularization parameter that was used to train the network, $\DS=\{\xv_i, y_i\}_{i=1,\dots,n}$ - training dataset, $k$ -- rank of the approximation}
    \Output{$D_{\Sigma^{-1}}, V$ - low-rank approximation to the parameter covariance matrix, $p^*$ - estimate of the effective number of parameters  }
     $B \leftarrow 0^{2k \times m}$ \;
     $\lambda_0 \leftarrow 0$ \;
     \For{$(\xv_i, y_i)$ in $\DS$}{
      $\mathbf{g}_i \leftarrow \nabla f_i = \left.\frac{\partial f}{\partial w} \right\vert_{x=x_i, w=\what}$ \;
      $B_i, \lambda_i \leftarrow$ \UpdateApproximation{$B, \lambda, \lambda_{i-1}, \mathbf{g}_i$} \; 
     }
     $U, D, V \leftarrow$ \SVD{$B_n$} \;
     $D_{\Sigma^{-1}} \leftarrow (D^2 + \lambda_n)(D^2 + (\lambda_n + \lambda) I)^{-2}$ \;
     $D_H \leftarrow (D^2 + \lambda_n)(D^2 + (\lambda_n + \lambda) I)^{-1}$ \;
     $p^* \leftarrow \textrm{Tr}(2D_H - D_H^2)$ \;
     \Return {$D_{\Sigma^{-1}}, V, \; p^* $}
     \caption{Estimate the covariance matrix of the NN parameters and effective number of parameters}
  \end{algorithm}

\subsection{Motivation behind the approximation}
  The proposed modification of the low-rank approximation procedure introduces the score function~\eqref{eqn:score_func} in order to capture the most important eigenvalues and eigenvectors of the parameter covariance matrix. We demonstrate the effectiveness of this heuristic on small-scale regression datasets from the UCI repository~\cite{UCI2019}. For a sufficiently small neural network it is feasible to apply formula~\eqref{eq2} directly. 

  \begin{figure}
    \centering
    \includegraphics[width=0.7\textwidth]{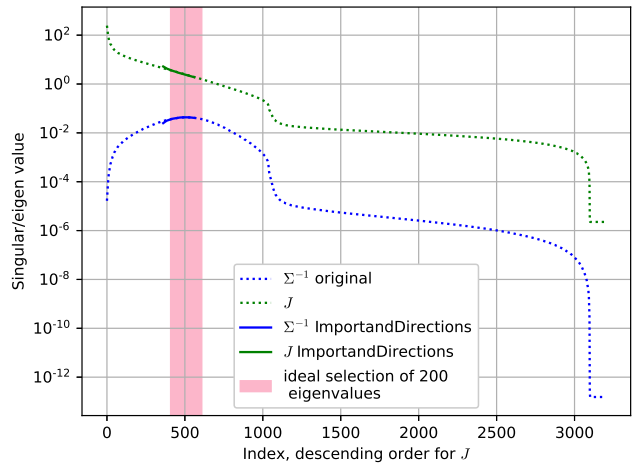}
    \caption{Approximation quality of Important Directions (rank $k=200$) for a $[50, 50]$ fully-connected network trained on the Wine Quality dataset. Dotted lines show the true specters of the matrices and solid lines show their respective approximations, shifted horizontally for visual clarity.}
    \label{fig:specters_wine}
  \end{figure}

  Figure~\ref{fig:specters_wine} shows comparison of the Important Directions approximation with exact computations: singular values of the Jacobian matrix $J$ and eigenvalues of the covariance $\Sigma^{-1}$ are plotted on the same axes. Using Robust Frequent Directions corresponds to selecting largest singular values of $J$, which will correspond to much smaller values for $\Sigma^{-1}$ (left-most part of the graph). At the same time, the largest eigenvalues of the covariance (shown in pink) will have more influence on the length of the resulting prediction intervals in~\eqref{eq2}. The presented graph shows that Important Directions approach achieves the goal of preserving the more informative part of the specter.

\subsection{Hyperparameters}
  The proposed method essentially has a single hyperparameter: the rank of the approximation $k$. Our experiments have shown diminishing returns with its increase.
  
  One advantage of this hyperparameter is that it can be chosen without the use of a hold-out dataset: based on the relative magnitude of the recovered singular values of the Jacobian. If the smallest and largest singular values of the approximation differ by several orders, additional singular values and singular vectors will contribute very little to the final length of the interval. An example of a sharp decline like this can be seen in Figure~\ref{fig:specters_wine} around index $1000$. 

  Since the size of the additional memory required for the approximation is a multiple of the total number of parameters of the neural network, it is reasonable to expect significant memory constraints playing a bigger role in the choice of this parameter. Thus we recommend to set it as high as your computational budget allows if you do not observe the behavior described in the previous paragraph.

\section{Experiments}
\label{sec:experiments}
  We evaluate our method against a number of existing approaches on publicly available regression datasets from UCI repository~\cite{UCI2019}. For comparison we have chosen methods designed specifically for neural networks and compared performance of the resulting prediction intervals.

  Every dataset was split 20 times into training and test sets and each method was independently applied on each split. Average performance metrics of these 20 runs are reported.

\subsection{Methods}
  \textbf{   MC-dropout.} We have based our implementation on the reference implementation\footnote{\url{https://github.com/yaringal/DropoutUncertaintyExps}} of~\cite{gal2016dropout}. Dropout probability is tuned via grid search on a validation set (taken out of training set) and the then the final model is trained from scratch on the whole training set. Resulting prediction interval is obtained using 10~000 stochastic forward passes.

  \textbf{Deep Ensembles.} We train the neural network multiple times from different initializations. Probabilistic output is a Gaussian with mean and variance obtained by aggregating their predictions.

  \textbf{Important Directions.} For our method we set rank of the approximation to a fixed value of 500 for most datasets with some exceptions: 200 for ``Wine'' and 300 for ``Boston'' due to their small size and 1000 for ``Year''. Number of training epochs is tuned on a validation set similar and then the network is retrained from scratch (similar to Dropout).

  For all methods we have used the training of a neural network as a black-box building block. The same fully connected neural network with two hidden layers of size 50 was used. We used Adam optimizer~\cite{kingma2014adam} with the default learning rate of $0.001$.

  \textbf{Neural Network Prediction intervals}~\cite{Veaux1998PI}. We have implemented the base method using the fact that our neural networks and some datasets are rather small. All the data was stored in GPU memory to speed up gradient computation. Jacobian product matrix $J^TJ$ was also pre-computed on GPU which will be impossible for larger networks.

\subsection{Metrics and results}
  While many existing papers on probabilistic regression use log-likelihood on the test set as the target metric, the quality of the resulting interval estimates is rarely explored~\cite{kompa2021empirical}. We argue that for a practitioner working on a regression problem a prediction interval with a specified level is much more useful than a simple ranking of the test sample by an uncertainty estimate. With this in mind we have chosen the following metrics to measure the quality of the obtained prediction intervals:
  \begin{itemize}
    \item Probability of coverage $p_{cov}$ -- proportion of the test examples for which estimated intervals cover the true response values.
    \item Pearson's correlation coefficient $r$ between width of the estimated interval and true absolute error.
    \item Width of the prediction interval in terms of the standard deviation of the true response values $w_{sd}$. Naturally one would prefer narrower intervals while still providing the coverage guaranties.
  \end{itemize}

\subsubsection{Comparison with the exact method.}
  We have compared our method with the exact application of formula~\eqref{eq2}, setting the rank at 500 for all of the datasets using a powerful workstation. While the computation time was impractical even for these simple datasets, it will allow us to assess whether our approximation is adequate. Results are summarised in Table~\ref{tab:exact}. Column $\Delta t$ corresponds to wall clock time in seconds that it took to run the method (excluding training of the neural network, etc.).
 
  As we can see the results for Important Directions are comparable with the exact computation. Rank of the approximation of 500 was enough even for larger datasets where $p^*$ is underestimated. Our hypothesis is that the original formula often produces over-conservative intervals -- their coverage is often greater than the nominal $95\%$, so the lower $p^*$ actually improves the metrics. See also the relevant discussion in~\cite{Veaux1998PI}.
 
  In some cases running time of the exact method is slightly better. This might be due to different implementations of SVD on CPU and GPU: we could not fit $J$ and/or $\Sigma^{-1}$ in the GPU memory.
  
  \begin{table}[t]
    \centering
    \caption{Comparison of the low-rank approximation with the exact method}\label{tab:exact}
      \begin{tabular}{lcccccccc}
        \toprule
        Metric & \multicolumn{2}{c}{$p_{cov}$} & \multicolumn{2}{c}{$r$} & \multicolumn{2}{c}{$w_{sd}$} & \multicolumn{2}{c}{$\Delta t$} \\
        Method &     Exact &    ID & Exact &    ID &    Exact &    ID &         Exact &   ID \\
        Dataset      &           &       &       &       &          &       &               &      \\
        \midrule
        Ailerons     &      0.94 &  0.95 &  0.22 &  0.25 &     1.55 &  1.55 &            25 &    6 \\
        Boston       &      0.98 &  0.96 &  0.36 &  0.34 &     1.61 &  1.33 &             7 &    1 \\
        CT           &      0.96 &  0.96 &  0.36 &  0.33 &     0.38 &  0.37 &          2398 &   24 \\
        Concrete     &      0.96 &  0.96 &  0.21 &  0.21 &     1.41 &  1.37 &             5 &    1 \\
        Energy       &      0.96 &  0.96 &  0.42 &  0.42 &     1.75 &  1.75 &             6 &    1 \\
        Protein      &      0.95 &  0.95 &  0.10 &  0.10 &     2.95 &  2.95 &            17 &   19 \\
        SGEMM        &      0.96 &  0.96 &  0.41 &  0.41 &     0.69 &  0.69 &            71 &   99 \\
        Superconduct &      0.94 &  0.94 &  0.10 &  0.12 &     1.50 &  1.52 &            76 &    9 \\
        Wine         &      0.98 &  0.95 &  0.18 &  0.14 &     3.85 &  3.08 &             6 &    1 \\
        Yacht        &      0.98 &  0.95 &  0.17 &  0.13 &     3.86 &  3.20 &             6 &    1 \\
        Year         &      0.94 &  0.93 &  0.22 &  0.22 &     3.10 &  3.07 &  5357 &  214 \\
        \bottomrule
      \end{tabular}
  \end{table}

\subsubsection{Comparison with other methods.}
  We have also preformed a comparison with other popular methods of uncertainty estimation for neural networks: MC-dropout and Deep Ensembles. Results of these experiments are presented in Table~\ref{tab:dropout}. For $p_{cov}$ the closest value to the nominal $95\%$ coverage is shown in bold. For $r$ the highest correlation coefficient is shown in bold.

  Overall it is a tie between ensembles and Important Directions: both methods are very close to the nominal coverage probability and provide intervals that correlate with the true error. Interesting observation is that methods are close in the three metrics at the same time, so ensemble does not for example provide universally tighter intervals.

  Our proposed explanation for the poor performance of the dropout is the following. Authors of~\cite{gal2016dropout} propose to tune both drop-out rate and $\tau$ parameter (related to noise precision) at the same time using grid search over predefined ranges. Even with standardizing the datasets it proves to be a challenging scenario given the observed coverage statistics.

  Given all the above we conclude that Important Directions method achieves results on par with more established but computationally heavy methods without the need to retrain the network and tune multiple hyper-parameters.

  \begin{table}[t]
    \centering
    \caption{Probability of coverage and other metrics for different methods.}\label{tab:dropout}
    \begin{tabular}{lccccccccc}
      \toprule
      Metric & \multicolumn{3}{c}{$p_{cov}$} & \multicolumn{3}{c}{$r$} & \multicolumn{3}{c}{$w_{sd}$} \\
      Method &         D-out &        Ens. &              ID & D-out &        Ens. &              ID &  D-out & Ens. &     ID \\
      Dataset      &                 &                 &                 &         &                 &                 &          &          &        \\
      \midrule
      Ailerons     &           0.965 &  \textbf{0.948} &           0.945 &   0.008 &           0.158 &  \textbf{0.245} &    1.960 &    1.573 &  1.550 \\
      Boston       &  \textbf{0.947} &           0.957 &           0.962 &   0.032 &           0.261 &  \textbf{0.344} &    1.519 &    1.388 &  1.330 \\
      CT           &           0.999 &           0.978 &  \textbf{0.956} &   0.005 &  \textbf{0.372} &           0.331 &    0.657 &    0.411 &  0.368 \\
      Concrete     &           0.908 &  \textbf{0.950} &           0.960 &  -0.062 &  \textbf{0.211} &           0.206 &    1.029 &    1.265 &  1.372 \\
      Energy       &           0.963 &  \textbf{0.952} &           0.959 &   0.022 &           0.419 &  \textbf{0.421} &    1.862 &    1.715 &  1.748 \\
      Protein      &           0.870 &           0.953 &  \textbf{0.952} &  -0.086 &           0.072 &  \textbf{0.097} &    1.960 &    2.965 &  2.949 \\
      SGEMM        &           0.990 &           0.967 &  \textbf{0.963} &  -0.069 &  \textbf{0.503} &           0.411 &    0.392 &    0.709 &  0.693 \\
      Sup.con. &           0.903 &  \textbf{0.944} &           0.943 &  -0.097 &  \textbf{0.271} &           0.121 &    1.029 &    1.523 &  1.524 \\
      Wine         &           0.801 &  \textbf{0.946} &  \textbf{0.946} &   0.032 &           0.114 &  \textbf{0.144} &    1.960 &    3.135 &  3.082 \\
      Yacht        &           0.802 &           0.946 &  \textbf{0.952} &   0.041 &  \textbf{0.139} &           0.133 &    1.960 &    3.135 &  3.202 \\
      Year         &           0.847 &  \textbf{0.936} &           0.934 &   0.006 &           0.186 &  \textbf{0.216} &    1.960 &    3.116 &  3.070 \\
      \bottomrule
    \end{tabular}
  \end{table}

\subsection{Implementation}
  We have implemented our method in Python 3 language. Neural network training and gradient computations were done in PyTorch~\cite{Paszke2019}. Subsequent matrix and vector calculations used NumPy library~\cite{harris2020array}.  Full source code for the method and experiments is made publicly available \footnote{\url{https://github.com/stat-ml/id_prediction_intervals}}.

\section{Discussion and future work}
\label{sec:discussion}
  While we introduce Important Directions as method for regression tasks, it can also be applied in the classification setting. For a given class index the network's logits can be combined in a one-vs-rest fashion to produce a single logit value. For this we will need to run our method independently for each class (although some gradient computations could be reused). Other ways to extend our method to classification task is the topic for future research.

  The described increase in computational cost urges us to find other ways to improve performance. One way is to limit the number of the parameters of the neural network that are considered in the gradient computation and related matrix operations. We can restrict our focus only to top layers or any other subset. This is similar in spirit to performing Bayesian inference on the last layer only. A direction of possible future research is the choice of higher rank vs larger number of parameters.

\section{Conclusions}
\label{sec:conclusions}
  This work presents a new method to construct prediction intervals for a trained neural network based on matrix sketching. The method was developed for regression tasks but can also be applied in a classification setting targeting class logits. We demonstrate its advantages compared to other interval estimation methods for neural networks on a range of benchmark regression datasets.

\noindent \textbf{Acknowledgements.} The research was supported by the Russian Science Foundation grant 20-71-10135.

\bibliographystyle{splncs04}
\bibliography{references}

\end{document}